\documentclass[letterpaper]{article} 
\usepackage{aaai25}  
\usepackage{times}  
\usepackage{helvet}  
\usepackage{courier}  
\usepackage[hyphens]{url}  
\usepackage{graphicx} 
\usepackage{natbib}  
\usepackage{caption} 
\usepackage{algorithm}
\usepackage{algorithmic}
\usepackage{amsmath}
\usepackage{makecell}

\usepackage{newfloat}
\usepackage{listings}
\usepackage{fancyvrb}
\usepackage{tcolorbox}
\DeclareCaptionStyle{ruled}{labelfont=normalfont,labelsep=colon,strut=off} 
\floatstyle{ruled}
\newfloat{listing}{tb}{lst}{}
\floatname{listing}{Listing}
\title{Skill-LLM: Repurposing General-Purpose LLMs for Skill Extraction}
\author {
    Amirhossein Herandi,
    Yitao Li,
    Zhanlin Liu,
    Ximin Hu,
		Xiao Cai
}
\affiliations {
    AstrumU Inc \\
    amir@astrumu.com, yitao@astrumu.com, kevin.liu@astrumu.com, ximin.hu@astrumu.com, xiao@astrumu.com
}

\usepackage{bibentry}

\begin{document}

\maketitle

\begin{abstract}
Accurate skill extraction from job descriptions is crucial in the hiring process but remains challenging. Named Entity Recognition (NER) is a common approach used to address this issue. With the demonstrated success of large language models (LLMs) in various NLP tasks, including NER, we propose fine-tuning a specialized Skill-LLM and a light weight model to improve the precision and quality of skill extraction. In our study, we evaluated the fine-tuned Skill-LLM and the light weight model using a benchmark dataset and compared its performance against state-of-the-art (SOTA) methods. Our results show that this approach outperforms existing SOTA techniques.
\end{abstract}

%
\begin{links}
    \link{Code}{https://github.com/herandy/Skill-LLM}
\end{links}

\section{Introduction}

The need for skills comprehension is always a demanding requirement regardless of the industry the job role belongs to. With the ever changing nature of jobs in the professional world and highly technical skills and domains, the need for algorithms for thorough understanding of relevant documents like job postings is ever present. It is important to be able to accurately and efficiently identify and extract applicable entities, such as skills, from these related resources. Aside from these methodologies, it is shown \cite{zhang2022skillspanhardsoftskill} that there is a need for datasets that contain more complex and descriptive entities or even technical acronyms (e.g. AWS, CI/CD) within relevant context to enable the state-of-the-art algorithms to embed all of these more novel skills and phrases alongside their relations.

The highly technical nature of skill related documents and domain specific phrases used in them means that smaller models with limited vocabularies and ones that may not have encountered in publicly available data are naturally not quite able to embed the most correct understanding of them during training and inference. It has been demonstrated that an initial unsupervised pretraining on job role related datasets \cite{zhang2022skillspanhardsoftskill, Zhang2023ESCOXLMRMT} containing sentences that include many of the domain specific phrases and skills, before the main supervised training, can improve skill extraction quality to some extent.

Recent approaches have leveraged large language models (LLMs) to address the task of skill extraction \cite{nguyen-etal-2024-rethinking, li2023skillgptrestfulapiservice}. These LLM-dependent methods eliminate the need for large tagged datasets; however, they possess some issues. One key issue is the requirement for specific and lengthy prompts to clearly define skill entities. Additionally, as highlighted by \cite{nguyen-etal-2024-rethinking}, pretrained LLMs often fail to produce results comparable to supervised methodologies. To mitigate this, techniques such as using few-shot examples from a tagged training set, either randomly selected or similar to the input, have been proposed. Another challenge with LLMs is that, as generative models, their outputs often require parsing for further use and evaluation. Depending on the quality of the output, some results may not be parsable without manual or automated modifications. Our study demonstrates that, with proper training, our method generates outputs that are consistently parsable. By fine-tuning LLMs, we achieve F1 scores that surpass those of both traditional supervised methods and LLM-based zero-shot and few-shot approaches.

Other works have expanded beyond the sole focus on the extraction task. \cite{Gugnani_Misra_2020} propose not only extracting skills that are explicitly mentioned in documents but also to identify implicit skills - those present in documents that are semantically similar to the original document. This approach is particularly relevant when the inputs are entire documents, such as job descriptions, rather than individual sentences. Their work also addresses methods and challenges related to matching skills between job postings and resumes. Similarly, to aid in the consolidation of skills that may be expressed in different forms (e.g. Python, Python programming language, Python coding), \cite{Zhao_Javed_Jacob_McNair_2015, Hoang_Mahoney_Javed_McNair_2018} focus on skill extraction and normalization. They introduce frameworks for generating taxonomies, which are then used to semantically match skills from new resources to the generated taxonomy database for the normalization task.

In this paper, we present a methodology that integrates the strengths of supervised learning with the broad generalization capabilities and extensive vocabulary of LLMs. Our approach demonstrates that by utilizing LLMs, we can bypass the need for additional pretraining on domain-specific datasets. This advantage is largely due to the substantial pretraining that LLMs undergo, which involves datasets of unprecedented scale. For instance, LLaMA 2 \cite{Touvron2023Llama2O} and LLaMA 3 \cite{dubey2024llama3herdmodels} have been pretrained on 2 trillion and 15 trillion tokens, respectively, in stark contrast to BERT’s \cite{Devlin2019BERTPO} and GPT-3’s \cite{brown2020languagemodelsfewshotlearners} pretraining on 3.3 billion and 300 billion tokens. The vast size of these pretraining datasets partially eliminates the necessity for further pretraining on specialized domain datasets.

Furthermore, by integrating this benefit with the proven capability of supervised training for the NER task, we can also avoid the requirement for complex prompts tailored to specific data and entity types. This synergy between supervised learning and LLM-based techniques allows us to leverage the strengths of both approaches, resulting in performance that surpasses current state-of-the-art methods.

In situations where computational resources are limited, we also offer a fine-tuning approach for a lightweight model that performs comparably to much larger baseline techniques for skill extraction.

\section{Related Work}

Skill extraction, a crucial component of workforce assessment, has evolved through various methodologies, ranging from manual tagging to the use of rule-based systems. The progression of skill extraction techniques has been significantly influenced by advancements in natural language processing (NLP). Early methods heavily relied on keyword matching, which limited scalability and adaptability, often resulting in oversimplifications and missed nuances.

The introduction of advanced techniques such as machine learning, particularly deep learning shines a light on solving the problem; however, challenges in accuracy and context understanding persisted. Recent advances in NLP, especially in transformer-based models, have led to the development of more advanced approaches for skill extraction. These innovations have provided more accurate and scalable solutions, addressing challenges that traditional methods could not overcome.

\subsection{Prompt Tuned LLMs}

Several methodologies have been developed to harness the power of LLMs for skill extraction, such as the work by \cite{Wang2023GPTNERNE}, which, while approaching the performance of supervised methods, still falls short in most cases. These LLM-based methods primarily rely on zero-shot and few-shot techniques. \cite{nguyen-etal-2024-rethinking} introduce two major output types for skill extraction: NER-style and extraction-style, with the latter consistently outperforming the former. To address parsing issues that can arise with NER-style outputs from LLMs, they propose heuristic techniques to correct these errors within the LLM itself. Despite these efforts, even advanced models like GPT-4, when used in a zero-shot or few-shot manner, do not achieve results comparable to those of supervised methods. \cite{nguyen-etal-2024-rethinking} suggest techniques to mitigate this gap, such as incorporating few-shot examples from tagged training sets, either selected randomly or chosen based on semantic similarity to the input.

One of the primary challenges with current methodologies based on prompt-tuned LLMs is the occurrence of hallucinations, where the model generates information that is not present in the input data. This can lead to significant inaccuracies in skill extraction tasks, as the model might produce results that do not fully align with the actual content. Additionally, ensuring consistency across different instances of extraction is a persistent issue. Even slight variations in prompts can lead to unpredictable and inconsistent outputs, which undermines the reliability of these methods.

Another significant concern is the format of the output. Prompt-tuned LLMs can struggle to maintain a consistent and reliable output format, especially when dealing with complex or nuanced data. This inconsistency can complicate the downstream processes, making it challenging to streamline the integration of skill extraction into larger systems or databases.

Together, these challenges highlight the limitations of relying solely on prompt-tuned LLMs for skill extraction. Addressing these issues is crucial for improving the accuracy, consistency, and reliability of the outputs, which are essential for the effective application of Named Entity Recognition.

\subsection{Fine Tuned models}

Current state-of-the-art models for skill extraction primarily consist of NER-based transformer models, often enhanced with additional techniques to improve extraction quality. \cite{zhang2022skillspanhardsoftskill} proposes JobBERT, demonstrating that pretraining models on untagged job posting data can lead to improved results. Another pretraining methodology, proposed by \cite{Zhang2023ESCOXLMRMT}, is ESCOXLM-R, which shows that using a multilingual dataset from the European Skills, Competences, Qualifications, and Occupations classification dataset (ESCO) \cite{6928765} for pretraining, combined with a multilingual model, can further enhance performance. Additionally, there are methodologies based on retrieval-augmentation using language models. For instance, \cite{zhang-etal-2024-nnose} suggest incorporating other skill extraction datasets as external data sources for retrieval to identify neighboring skills, thereby improving extraction accuracy.

The technical language found in documents like job postings presents challenges for current skill extraction methods. These documents often use specialized terminology that may not be well-represented in general-purpose datasets, making it essential for base models to undergo further pretraining on domain-specific data. For traditional models, this means additional pretraining on relevant datasets to ensure accurate skill extraction. Similarly, LLMs should ideally be pretrained on the most current data available, encompassing a wide range of technical terms from various fields.

\section{Methodology}

This paper addresses the challenge of extracting skills from job postings, a task that is further complicated by the semantic diversity across different industries and job roles. Accurately identifying and extracting these skills is crucial for understanding industry requirements and enhancing the efficiency of recruitment processes.

\subsection{Fine Tuning LLMs and Limitations}

\paragraph{Prompt Construction}
Named Entity Recognition (NER) tasks are highly contextual and dependent on grammar. Recent studies have explored the potential of leveraging large language models (LLMs) for skill extraction, particularly in understanding and extracting skills from unstructured text. Our work extends this line of inquiry by repurposing general-purpose LLMs, thereby enhancing the accuracy and adaptability of skill extraction. By leveraging the reasoning power of LLMs alongside their vast vocabulary and extensive number of trainable parameters, we aim to improve the quality of NER tasks. Since LLMs are primarily generative in nature, we have adapted them for NER tasks by fine-tuning them to generate outputs as JSON-formatted dictionaries. To further improve the position matching of extracted entities, especially in cases where the same entity appears more than once in a sentence with different meanings (e.g. \textit{"The job requires work in the office and the ability to operate office equipment"}), we propose the following format:

\begin{figure}[ht]
\begin{tcolorbox}[colback=green!6, colframe=black]
\textbf{Prompt}:

\lstinline|You are a helpful information extraction system. Your job is to extract skill entities and knowledge entities from the given sentence.|\\

\textbf{Query}:

\lstinline|** You will be working in an end-to-end cross-functional team being responsible for implementing and promoting all QA relevant topics on team level . **|\\

\textbf{Response}:

\begin{lstlisting}
{"SKILL": [
  {"skill_span": "implementing and promoting all QA relevant topics",
   "context": "for implementing and promoting all QA relevant topics on"}
 ],
 "KNOWLEDGE": [
  {"skill_span": "QA",
   "context": "all QA relevant"}
 ]
}
\end{lstlisting}
\end{tcolorbox}
\caption{Example for fine tuned LLM input and output formats}
\label{fig 2: prompt}
\end{figure}

As shown in Figure \ref{fig 2: prompt}, each type of entity is represented as a list of dictionary elements, with each element comprising two key components: ``skill\textunderscore span'' and ``context''. The ``skill\textunderscore span'' refers to the specific entity extracted from the text, while the context includes the ``skill\textunderscore span'' itself, along with one additional token from the text on either side. This method ensures that the extracted entity is surrounded by its immediate textual environment, providing essential contextual information.

To handle situations where the ``skill\textunderscore span'' occurs at the beginning or end of the input, we introduce special tokens on either side of every input sequence before passing it to the model. We have selected the ``**'' symbol as this special token to ensure consistency in context extraction regardless of the position of the ``skill\textunderscore span'' within the text. This approach standardizes the input format, enabling the model to process and extract entities effectively across various input scenarios.

One of the main advantages for this output format is that it allows us to map the entities back to the original text accurately. Most evaluation metric calculations for NER tasks will be reliant on the position of the extracted entities from the text and the context will allow us to do the evaluation correctly. This format also adds an additional layer of reliability assurance, as we can verify that the mapping is done precisely every time, especially when there exist multiple instances of the same exact word or phrase appearing in the input text.  For example, in the sentence \textit{"The job requires work in the office and the ability to operate office equipment"}, the first instance of ``office'' would ideally not be recognized as a skill, while the second instance would be correctly identified. The matching of the skill span, context, and input text ensures that the generated output is accurate and minimizes the chances of hallucination by the LLM in use.

By aligning the skill span, the surrounding context, and the original text, we can enhance the accuracy of the generated output. This alignment strategy also allows us to detect errors when they appear, such as hallucinations, which can occur when the language model generates content that deviates from the intended or factual output. Therefore, our approach not only ensures correctness but also reinforces the overall reliability of the system in handling complex input texts.

A key benefit of utilizing a fine-tuned LLM model in our approach is that it removes the necessity for lengthy and complex prompts. Typically, such prompts are designed to specify different entities and enforce a rigid output format, often accompanied by cautionary instructions. In contrast, as illustrated in Figure \ref{fig 2: prompt}, our fine-tuned LLM model is simply instructed to extract all skill and knowledge entities from the input sentence provided. This streamlined approach not only eliminates the need for intricate prompt engineering but also minimizes the total number of tokens processed by the model with each call.

For a comprehensive comparison, we have included examples of equivalent input and output sequences using different methodologies. Figures \ref{fig 3: extract style prompt} and \ref{fig 4: ner style prompt} illustrate the Extract-Style and NER-Style methods, respectively, which were previously discussed by \cite{nguyen-etal-2024-rethinking}. These figures provide a visual contrast to our approach, highlighting the differences in prompt design and model outputs.

\begin{figure}[ht]
\begin{tcolorbox}[colback=green!6, colframe=black]
\textbf{Prompt}:

\lstinline|You are an expert human resource manager.|

\lstinline|You need to analyse skills required in job offers. You are given a sentence from a job posting. Extract all the <SKILL TYPE> that are required from the candidate as list, with one skill per line.|\\

\textbf{Query}:

\lstinline|You will be working in an end-to-end cross-functional team being responsible for implementing and promoting all QA relevant topics on team level .|\\

\textbf{Response for skill}:

\lstinline|implementing and promoting all QA relevant topics|\\

\textbf{Response for knowledge}:

\lstinline|QA|\\

\end{tcolorbox}
\caption{Example for Extract-Style LLM prompting}
\label{fig 3: extract style prompt}
\end{figure}

\paragraph{Limitations}

Similarly to other LLM-based models, the generated outputs need to be parsed before they can be used for evaluation. In rare cases, some outputs may be unparsable due to issues such as incorrectly formatted JSON strings (e.g. malformed braces and brackets) or the model exceeding its available context length for the output. Some works, such as \cite{nguyen-etal-2024-rethinking}, attempted to heuristically address this issue by modifying the prompt. However, this approach is not feasible for our methodology due to practical constraints during the fine-tuning process. Therefore, our focus is on fine-tuning the model itself to mitigate these issues effectively.

One of the primary limitations of LLM-based methodologies lies in the computational demands of LLM deployment. Due to their substantial size and complexity, LLMs require significantly more computational resources than traditional NER methods. This increased demand for processing power directly impacts inference time, making it considerably slower compared to smaller transformer-based models commonly used in NER tasks. As a result, while LLMs may offer improved accuracy and flexibility, the trade-off in terms of speed and efficiency is a notable drawback in practical applications.

\begin{figure}[ht]
\begin{tcolorbox}[colback=green!6, colframe=black]
\textbf{Prompt}:

\lstinline|You are given a sentence from a job posting.|

\lstinline|Highlight all the <SKILL TYPE> that are required from the candidate, by surrounding them with tags '@@' and '##'. If there are no such element in the sentence, replicate the sentence identically.|\\

\textbf{Query}:

\lstinline|You will be working in an end-to-end cross-functional team being responsible for implementing and promoting all QA relevant topics on team level .|\\

\textbf{Response for skill}:

\lstinline|You will be working in an end-to-end cross-functional team being responsible for @@implementing and promoting all QA relevant topics## topics on team level .|\\

\textbf{Response for knowledge}:

\lstinline|You will be working in an end-to-end cross-functional team being responsible for implementing and promoting all @@QA## relevant topics on team level .|\\

\end{tcolorbox}
\caption{Example for NER-Style LLM prompting}
\label{fig 4: ner style prompt}
\end{figure}

\subsection{GLiNER}

The previous method leverages a large model for accuracy. However, for portability, we propose an alternative approach utilizing a lightweight model. To ensure compatibility with this model, the training data must be preprocessed into a specific format. Specifically, GLiNER \cite{zaratiana2023glinergeneralistmodelnamed} requires the input data to consist of a list of ``tokenized\textunderscore text`` along with a corresponding list labeled ``ner``. The ``ner`` list contains elements, each defined by a start index, an end index, and the associated entity label, as illustrated in Figure \ref{fig 5: GLiNER data format}.

\begin{figure}[htbp]
\begin{tcolorbox}[colback=green!6, colframe=black]
\textbf{tokenized\textunderscore text}:

\lstinline|['You','will','be','working','in','an','end-to-end','cross-functional','team','being','responsible','for','implementing','and','promoting','all','QA','relevant','topics','on','team','level','.']|

\textbf{ner}:

\lstinline|[[12, 18, 'Skill'], [16, 16, 'Knowledge']]|

\end{tcolorbox}
\caption{Example for GLiNER data format}
\label{fig 5: GLiNER data format}
\end{figure}

Although GLiNER can be applied as a zero-shot method, we chose to fine-tune the model to take of advantage of the same benefits as other supervised approaches.






\section{Experiments}

\subsection{Dataset}

For the purposes of this paper, we will be focusing on the SkillSpan dataset, which has been curated specifically for the task of skill extraction by \cite{zhang2022skillspanhardsoftskill}. This dataset includes a diverse range of job postings across multiple industries, providing a robust foundation for evaluating the effectiveness of our proposed methods. The SkillSpan dataset is particularly valuable because it captures both explicit and implicit skills, allowing for a comprehensive assessment of model performance in real-world scenarios. The data is split into three different sets: train, validation, and test. State-of-the-art methodologies have been selected based on validation data metrics, with results on the test sets reported and compared in various papers \cite{zhang2022skillspanhardsoftskill, Zhang2023ESCOXLMRMT, nguyen-etal-2024-rethinking, zhang-etal-2024-nnose}. Given the dataset's prior use in related studies, it serves as a reliable benchmark, allowing for direct comparison across different methods. The dataset was originally compiled from job postings from three different sources: big, house, and tech. The tech portion primarily consists of sentences from job postings taken from job posting of StackOverflow platform, while the house and big portions contain sentences from a variety of job postings across different industries and sectors. The publicly available dataset consists of the house and tech portions, which are split into train, validation, and test sets. The numbers for each entity per set are shown in Table \ref{tab 1: dataset}.

\begin{table}[htbp]
\centering
\begin{tabular}{|l|c|c|c|}
\hline
\textbf{Entity Type} & \textbf{Train} & \textbf{Validation} & \textbf{Test}\\
\hline
Knowledge & 2969 & 1093 & 1174 \\
Skill & 2221 & 1070 & 1091 \\
\hline
\end{tabular}
\caption{Entity count breakdown for the \textbf{SkillSPAN} dataset. This datasets consists of sentences extracted from job postings and includes spans tagged for skills and knowledge.}
\label{tab 1: dataset}
\end{table}

\subsection{Baseline methods}

To assess the effectiveness of our proposed approach, it is essential to establish baseline methods for comparison.

In our comparison, we have carefully selected and presented the best span F1 scores reported in each paper (See Table \ref{tab 2: results}). If the paper describes multiple approaches, such as incorporating additional pre-training or post-processing steps, we have included a subset of these methods to enhance the comparability of results. This selective inclusion ensures that our comparison remains rigorous and relevant, allowing for a more meaningful analysis of the various methodologies.

The first baseline method, introduced by \cite{zhang2022skillspanhardsoftskill}, is called \textbf{JobBERT}. This method introduces the dataset that we will be using for our comparison and shows different results based on various models and methodologies. Depending on the split of the dataset or entity type used, either JobBERT or JobSpanBERT can yield higher or lower span F1 scores. For the sake of comparison, we focus on the best results they reported on the test split, for the combination of the entity types, skills and knowledge. Both models are pretrained on an untagged dataset consisting of 3.2M sentences from 127K job postings. We also include results from models not pretrained on this additional dataset to isolate the effects of supervised methodology, highlighting the difference compared to a model like GLiNER without extra pretraining or post-processing. Due to its flexibility and strong performance across various NLP tasks, BERT \cite{Devlin2019BERTPO} is used as a token classification model. Similarly, SpanBERT \cite{Joshi2019SpanBERTIP} is trained to identify spans of tokens, focusing not only individual tokens but also span boundaries.

Next, we consider \textbf{ESCOXLM-R} \cite{Zhang2023ESCOXLMRMT}, which is based on the XLM-RoBERTa large model \cite{Conneau2019UnsupervisedCR} and utilizes a different dataset for pretraining. By using multilingual models and pretraining datasets, this method achieves higher scores compared to previous baselines. The database used is a multilingual entity linking dataset from ESCO, and the authors introduce two new pretraining objectives: one based on LinkBERT \cite{yasunaga-etal-2022-linkbert}, and the other on ESCO Relation Prediction. These objectives incorporate relationships and connections between different occupations, skills, and competencies during pretraining.

The third baseline method, introduced by \cite{nguyen-etal-2024-rethinking}, does not use fine-tuning. While previous studies indicate that LLMs often cannot compete with supervised methodologies for entity extraction \cite{Wang2023GPTNERNE, Ma2023LargeLM}, this work attempts to bridge the gap between prompt-based generative models and supervised ones. For the generation task, they compare two main formats for extraction and use few-shot samples from the training set to improve LLM extraction quality. They propose several methods for selecting these examples, including semi-random selection and kNN-retrieval. For our comparison, we report results from the Extract-Style format with semi-random selection, as it was the best-performing method on our dataset.

The fourth baseline, \textbf{NNOSE}, introduced by \cite{zhang-etal-2024-nnose}, similarly uses nearest neighbors for retrieval-augmentation. This method retrieves semantically similar text from multiple datasets to enhance skill extraction, using a more advanced transformer model (RoBERTa \cite{Liu2019RoBERTaAR}) for both training and inference. It's important to note that NNOSE flattens both entity types into one during data preprocessing, which might affect direct comparability with other results. The results presented here are based on the best-performing version, which uses whitening transformation and selects nearest neighbors from all the training datasets, including SkillSpan \cite{zhang2022skillspanhardsoftskill}, Sayfullina \cite{sayfullina2018learningrepresentationssoftskill}, and Green \cite{green-etal-2022-development}.


\setlength{\tabcolsep}{1mm}

\begin{table*}[ht]
\centering
\begin{tabular}{|l|c|c|c|c|c|c|}
\hline
\textbf{Method} & Parameters&\textbf{\thead{Skill\\Span F1}} & \textbf{\thead{Knowledge\\Span F1}} & \textbf{\thead{Total\\Span F1}} & \textbf{Training \& Inference}\\
\hline
BERT \cite{zhang2022skillspanhardsoftskill} & 110M & 54.2\% & 61.7\% & 57.7\% & fine tuning\\
jobSpanBERT \cite{zhang2022skillspanhardsoftskill} & 110M & \textbf{56.3\%} & 61.9\% & 58.9\% & pretrain + fine tuning \\
ESCOXLM-R \cite{Zhang2023ESCOXLMRMT} & 550M &- & - & 62.6\% & pretrain + fine tuning \\
GPT-3.5 NER-Style \cite{nguyen-etal-2024-rethinking} & 175B &- & - & 17.8\% & 5-shot \\
GPT-3.5 Extract-Style \cite{nguyen-etal-2024-rethinking} & 175B & - & - & 25.0\% & 5-shot \\
GPT-4 \cite{nguyen-etal-2024-rethinking} & 1.7T &- & - & 27.8\% & $\leq$ 350 subset \\
NNOSE \cite{zhang-etal-2024-nnose} & 123M &- & - & 64.2\% & pretrain + fine tuning + kNN \\
Fine-tuned GLiNER (Ours) & 166M &49.6\% & 65.5\% & 58.4\% & fine tuning \\
Skill-LLM - Finetuned LLaMA 3 8B (Ours) & 8B & 54.3\% & \textbf{74.2\%} & \textbf{64.8\%} & fine tuning \\
\hline
\end{tabular}
\caption{We compare Span F1 results for all entity types in the SkillSpan test dataset, skills and knowledge. The results highlight the performance of each model, they also demonstrate how training and inference strategies impact overall outcomes.}
\label{tab 2: results}
\end{table*}

\subsection{Experimental Setup}

For Large Language Model fine tuning, we have selected LLaMA 3 8B \cite{dubey2024llama3herdmodels} as the baseline. LLaMA-Factory \cite{zheng2024llamafactory} has been used for fine tuning and generation of evaluation outputs, which streamlines the fine tuning process and output generations for a multitude of publicly available large language models and their expected prompt formats. Low-Rank Adaptation (LoRA) \cite{hu2021loralowrankadaptationlarge} has been used to both speed up and reduce memory requirements during training. The generated outputs are then parsed into line delimited json format consisting of text, tokens, and spans that are merged with the gold label data for NER evaluation based on the methodology introduced in \cite{segura-bedmar-etal-2013-semeval}. The best model and hyperparameters are selected based on the best performing model on the validation set of the SkillSPAN dataset and then using that model, inference is run on the test set for the report.

Given the significant time investment required for training large language models (LLMs), we opted not to engage in traditional hyperparameter tuning methods. Instead, we employed an iterative approach, where parameters were adjusted based on performance metrics observed on the validation set. This approach allowed us to refine our model more efficiently.

The final hyper-parameters selected for our experiments are as follows. We utilized the LLaMA 3 8B Instruct model as the baseline model. The learning rate was set to 2e-4, and a batch size of 4 was chosen to ensure stability during training. To enhance the model's efficiency, we applied Low-Rank Adaptation (LoRA) with a rank of 64, targeting the query projection (q\textunderscore proj) and value projection (v\textunderscore proj) matrices. We conducted fine-tuning over 2 epochs, which was deemed sufficient based on the observed convergence of the model and the weight decay was set to 0. Additionally, the training process included a 10\% initial warmup period, followed by a cosine scheduler to dynamically adjust the learning rate.

Although LLMs are generally slower and more costly to run compared to smaller transformer models typically used for supervised learning tasks, our fine-tuned LLM method offers distinct advantages over traditional prompt-based LLM techniques.

One benefit is that, after the training phase is complete, our method requires a considerably shorter prompt during inference. Unlike traditional approaches that rely on detailed instructions regarding output format or definitions of entity types, our fine-tuned LLM has already internalized this information during the supervised training process. As a result, the model can generate accurate outputs with minimal input, improving both efficiency and ease of use. The specific prompt currently employed in our approach is illustrated in Figure \ref{fig 2: prompt}.

For the GLiNER model, the baseline model is the ``gliner\textunderscore small\textunderscore news-v2.1'', we employed Ray Tune \cite{liaw2018tuneresearchplatformdistributed} to conduct a comprehensive hyperparameter search and optimization process. Given that this model is significantly smaller in size compared to LLMs, we were able to explore a wide range of hyper-parameter configurations without the burden of excessive computational demands. Running the model on the dataset, we carefully selected the configuration that performed best on the validation set. After identifying the optimal model and checkpoint, we then evaluated its performance on the test set, and these results are reported in Table \ref{tab 2: results}. The learning rate for the pre-trained layers was set to 1.6e-4, while the non-pretrained layers, feedforward and span representation, were assigned a learning rate of 4e-5, a weight decay of 0.03 was applied and the maximum gradient norm  was capped at 0.4.

\section{Results}

\subsection{Quantitative Analysis}
The best F1 metric values are taken from their respective papers for comparison, in some cases, multiple models or methodologies are selected from each paper for better comparison based on training and testing methodologies. This metric focuses on the exact matching of the extracted spans and the precise boundaries for them. Each of these techniques possesses its own strengths and weaknesses. By analyzing their performance, we gain a better understanding of how different methodologies impact the final results. This allows us to draw better conclusions about the relative advantages of these models for skill extraction. Individual F1 results for skills and knowledge were only reported in cases where they were present in their respective papers as well as GLiNER and Skill-LLM where the experiments were conducted by us.

The evaluation was set up to ensure that the results would be both comparable and robust. By using the SkillSPAN dataset and focusing on the span F1 metric across all entity types, the analysis offers a comprehensive overview of the model's effectiveness. This approach allows for a clear comparison with existing literature, ensuring that the findings contribute meaningfully to the ongoing discourse in the field of skill extraction.

Our method outperforms current state-of-the-art models, achieving an F1 score of 64.8\% on the test set and 64.7\% on the validation set. This is accomplished without relying on additional pretraining or retrieval-augmented methods during inference. Notably, our approach demonstrates strong performance across different entity types. On the test set, our model achieves an F1 score of 54.3\% for skills and 74.2\% for knowledge. Similarly, on the validation set, it records 60.0\% for skills and 69.0\% for knowledge.

Our fine-tuned lightweight model also demonstrates strong performance despite having a relatively small number of parameters. Remarkably, it achieves results comparable to those of much larger models.

\subsection{Qualitative Analysis}

Errors can occur when the model's outputs do not exactly match the gold labels. Given the limited size of any dataset, it is inevitable that the model may not encounter every possible variation during training. Discrepancies between the training data distribution and test data, along with minor inconsistencies in tagging due to human error, can degrade the quality of the final output. Additionally, the nature of machine learning models, particularly those with high-dimensional parameters, often leads to the discovery of local optima rather than a true global minimum.

In this study, we identified several key challenges specific to our method and dataset. One major issue is the inconsistency of LLM outputs, where formatting errors rendered the results unparsable for further evaluation. Another challenge, as noted by \cite{nguyen-etal-2024-rethinking}, is the presence of problematic gold label annotations, which contribute to lower overall performance scores, 8\% of error cases are cause by this which are job market related entities that are not skill entities. For instance, in the sentence "DevOps Engineer ( CI CD Cloud Docker Jenkins ) $<$ORGANIZATION$>$", the gold labels were incorrectly left empty for both skills and knowledge. Similarly, in the sentence "Result oriented and work constructively towards reaching goals.", the model produced an output with a misplaced character, a ')' in place of a '\}', leading to parsing errors:


\begin{lstlisting}
{"SKILL":
  [{"skill_span": "Result oriented",
    "context": "** Result oriented and"),
   {"skill_span": "work constructively towards reaching goals",
    "context": "and work constructively towards reaching goals ."}],
 "KNOWLEDGE": []}}
\end{lstlisting}

These errors are really rare, only happening in one case out of 3174 inputs for the validation set and no error cases for the test set.


\subsection{Discussion}
It is important to highlight that the second-highest overall score on the test set, attributed to the NNOSE model, is achieved by merging both skills and knowledge into a single entity type within their pipeline. This approach may not offer a fair comparison as it simplifies the challenge by treating distinct entity types as one, potentially skewing the results in favor of their method. In contrast, our model maintains its superior performance while accurately distinguishing between skills and knowledge as separate entities.

The results presented align with our initial expectations, underscoring the effectiveness of our approach. LLMs are inherently powerful due to their extensive training on vast datasets and the large number of parameters they possess. This allows them to capture and integrate complex semantic relationships, particularly those involving specialized concepts like skills and knowledge, which may not frequently appear in everyday language. Our method leverages this inherent strength by applying a supervised learning approach, further refining the model's ability to discern and categorize these nuanced relationships effectively. The combined impact of these factors is reflected in the robust performance of our method, as demonstrated by the results detailed in Table \ref{tab 2: results}.

There are numerous practical applications for this method. Through extension, in recruitment, it could be used to quickly and accurately find candidates with specific skill sets, streamlining the hiring process. By automating skill extraction, organizations can ensure that they focus on the most relevant candidates, improving efficiency and decision-making in human resources.

\section{Conclusion and Future Work}
By fine-tuning large language models (LLMs) on tagged data and proposing a structured output format for skill extraction, we have introduced a method that enhances the quality and coverage of skill extraction while still leveraging the power of LLMs. Our fine-tuning approach takes advantage of the extensive vocabulary, strong generalization capabilities, and vast number of trainable parameters in LLMs, resulting in significantly improved performance compared to state-of-the-art methods, particularly for longer or more complex entities. Additionally, we introduced a methodology for fine-tuning a lightweight model that, despite its smaller size, achieves performance comparable to significantly larger models.

The success of this approach highlights the potential for adapting language models to specific tasks like skill extraction using existing datasets, without the need for novel ones. Future research could explore extending this methodology to other languages or adapting it for use in different industry sectors, provided the appropriate datasets are available. Moreover, given the potential for improved performance with increased training data, utilizing LLMs to generate synthetic data could be a promising direction for future work. This research contributes to the growing body of evidence supporting the use of natural language processing in enhancing human resource processes.
\bibliography{herandi}

\end{document}